\documentclass[5p,times]{elsarticle}
\usepackage{ecrc}

\volume{00}

\firstpage{1}

\journalname{Manufacturing Letters}

\runauth{N.D.K. Nguyen, H. Ko}

\jid{MFGLET}

\usepackage{amssymb}
\usepackage{amsmath}
\usepackage{booktabs}
\usepackage{xcolor}
\usepackage{algorithm}
\usepackage{algpseudocode}

\usepackage[figuresright]{rotating}

\usepackage[bookmarks=false]{hyperref}
    \hypersetup{colorlinks,
      linkcolor=blue,
      citecolor=blue,
      urlcolor=blue}

\begin{document}
\begin{frontmatter}

\dochead{55th SME North American Manufacturing Research Conference (NAMRC 55, 2027)}%

\title{Process-Knowledge-Guided Latent Diffusion for Generative Design of Porous Architectures: A Case Study in Aerosol Jet Printing}

\author[a]{Ngoc Dang Khoa Nguyen}
\author[a]{Hyunwoong Ko\corref{cor1}}
\ead{Hyunwoong.Ko@asu.edu}

\address[a]{School of Manufacturing Systems and Networks, Ira A. Fulton Schools of Engineering, Arizona State University, Mesa, AZ 85212, United States}

\begin{abstract}
Printed porous layouts interleave functional material with open channels, and the layout that uses the material best depends on how the process settings shape the deposit. Generative models propose layouts quickly, but a generator that does not receive the process setting samples from the same distribution at every setting, even when the process-dependent score ranks candidate layouts differently. This paper examines whether process knowledge can steer a latent diffusion generator toward layouts suited to a given process setting in a prescribed design domain. We formulate a process-dependent utilization score in which a process model sets the length over which the utility of printed material decays with distance from a channel. The generator combines a vector-quantized variational autoencoder (VQ-VAE) latent representation, a topology U-Net, a ControlNet domain branch, and a separately trained regressor that predicts the score from noised latents; the domain branch conditions each denoising step and the regressor gradient steers it. The method is applied to aerosol jet printing, where the atomizer and sheath gas flows enter the score through the flow-focusing relation. Across seven modeled flow settings, the lowest-scoring candidate layout of a domain differed between settings in 191 of 500 domains. In a sampling study, the mean directly computed score on three validation domains and one test domain was 0.642 with domain conditioning alone, 0.636 at a weight of 300, and 0.450 at 12,000 (112 scores per weight; lower is better). The results support process-knowledge-guided generation as an approach to process-specific porous design under the stated assumptions.
\end{abstract}

\cortext[cor1]{Corresponding author.}

\end{frontmatter}

\section{Introduction} \label{sec:intro}

Agentic artificial intelligence (AI) is entering engineering design and manufacturing. Agents based on large language models (LLMs) can use external tools to carry out a task and use the returned results to decide what to do next \cite{wang2024agents}. Such agents have planned and executed chemical experiments \cite{boiko2023}, revised truss designs using finite element feedback \cite{jadhav2024}, and adjusted material-extrusion printing parameters from in-process images \cite{jadhav2025print}. In engineering design, agents can use generative models to propose shapes or geometry parameters and use surrogate models to predict properties of the resulting designs \cite{elrefaie2025,zheng2026}. For an agent comparing design alternatives, a useful generator responds to specified design conditions and returns geometry that a separate evaluator can score. For porous architectures, this paper asks how a generator can respond to a specified process setting and how the resulting layout can be scored independently of the predictor used to guide generation.

Porous architectures deliver function through the arrangement of solid material and open pathways.\enlargethispage{-\baselineskip} Thick battery electrodes gain rate capability when aligned channels shorten the ion path through the electrode \cite{bae2013}, tissue scaffolds rely on interconnected pores for transport and cell ingrowth \cite{hollister2005}, and microchannel structures remove heat from microelectronic devices \cite{tuckerman1981}. In each case the internal pathways and the material distribution must be designed together with the limits of the fabrication process: a layout can fit inside its boundary and still contain a wall too thin to deposit or a channel too narrow to stay open. Containment within the design domain, geometric feasibility under a stated process window, and functional performance are therefore three different attributes of a layout.

Engineered porosity of this kind differs from the unintended porosity that degrades additively manufactured parts \cite{debroy2018}: it is specified by the designer, and its preferred amount and arrangement vary with the application. This paper considers planar porous layouts in which a printed region of functional material is interleaved with open channels inside a prescribed design domain, for processes whose settings shape the deposit. Aerosol jet printing (AJP), in which the atomizer and sheath gas flows shape the deposit \cite{secor2018,mahajan2013,wilkinson2019}, is the process of the case study.

Deep generative models have become a practical way to explore large design spaces \cite{regenwetter2022}. Once trained, a generative model can propose alternative layouts without solving a new iterative topology-optimization problem for each sample. Diffusion models \cite{ho2020} suit this role because the denoising process accepts conditions and gradient guidance during sampling \cite{dhariwal2021,ho2022cfg}, and latent diffusion models \cite{rombach2022} move the process to a compressed representation learned by an autoencoder. We adopt latent modeling as the representation in this work. A geometry that resembles the training data is only a starting point, however: the generator must respond to the engineering objective and to the conditions under which the design will be manufactured.

The specific challenge addressed here is to generate diverse porous layouts that remain inside a prescribed domain and that improve a functional objective whose value depends on the process setting. At the model level, this requires coordinating two kinds of information through one latent representation: spatial control from the domain and objective guidance from a process-dependent assessment. The guidance signal comes from a learned surrogate evaluated on noisy latents, so the sampler can exploit surrogate error, and the decoder can alter the fine features on which the objective depends. For this reason, the decoded layouts are evaluated directly under the prescribed process setting.

The proposed approach combines a latent representation derived from a vector-quantized variational autoencoder (VQ-VAE) \cite{oord2017}, a topology-generating U-Net, a ControlNet domain branch \cite{zhang2023controlnet}, and a separately trained regressor that predicts the process-dependent design score from the noisy latent, the domain latent, and the process setting. The domain branch conditions each denoising step, and the gradient of the regressor steers the sample toward lower scores. The guidance target is the utilization score, and containment within the domain is measured on the decoded layouts.

The contributions are threefold. First, we formulate a process-dependent utilization score for porous layouts, in which a process model sets the utilization length, together with a fixed-geometry audit that shows whether a change of process setting changes the lowest-scoring layout within a fixed candidate pool. Second, we integrate domain control and process-score guidance in a single latent sampler in which the process model supplies the training targets of the regressor and only the regressor receives the process setting. Third, in a case study on AJP, we apply the audit across seven settings of the atomizer and sheath flows and, with domain conditioning fixed, compare three guidance weights with zero guidance on ten domains, four of them held out from training, using scores computed directly on the decoded layouts. For agentic design workflows, the sampler can generate candidate layouts for a specified domain and process setting, and the evaluator computes the utilization score and domain error directly from each decoded layout without using the guidance regressor.

Section~\ref{sec:lit} reviews manufacturing-aware design of porous architectures and generative optimization and states the questions examined in this paper. Section~\ref{sec:method} formulates the method for a general process setting, with its training losses and evaluation measures, and Section~\ref{sec:case} applies it to AJP. Section~\ref{sec:discussion} discusses engineering implications and limitations, and Section~\ref{sec:conclusion} gives concluding remarks and future work.

\section{Literature review} \label{sec:lit}

\subsection{Design for Manufacturability of Porous Architectures} \label{sec:lit-dfm}

Manufacturability in topology optimization is commonly expressed through length-scale control: minimum solid and void feature sizes are imposed through filtering and projection \cite{lazarov2016,sigmund2007}, and additive manufacturing adds process-specific restrictions \cite{liu2018}. For porous functional structures, the same geometric features also govern performance. Electrodes with aligned channels at a second, larger porosity scale have lower effective tortuosity and higher usable capacity at high rates \cite{bae2013}, and porous electrodes have been designed by topology optimization \cite{roy2022,li2024}. Usseglio-Viretta et al. \cite{usseglio2023} optimized pore-channel patterns in battery electrodes with a genetic algorithm, using the average distance from the electrode material to the nearest channel as a proxy for fast charging and, with only the channels connected to the electrolyte entry edges counted, for wetting. The score used in this paper weights each printed pixel by a utility that decays with its distance from the nearest channel, over a length set by a process model (Section~\ref{sec:problem}). Minimum widths of the deposited features and of the open channels address whether a layout can be made; access of the channels from the boundary through which the transported species enters is a separate requirement.

The relation between process settings and the deposit is specific to each process. In AJP, the process of the case study, an annular sheath gas focuses the aerosol carried by the atomizer gas, and the ratio of the two flows controls the diameter of the focused stream relative to the nozzle \cite{secor2018,wilkinson2019}. Printed line width and thickness vary with both flows \cite{mahajan2013,phuah2020}, but converting the focused-stream diameter into a deposited feature width requires calibration for the ink, nozzle, and stage speed, and operating windows are usually identified empirically or with data-driven models \cite{zhang2019ajp}.

\nocite{maze2023,nie2021,giannone2023}
\begin{table*}[t]
\caption{Closest prior architectures, the corresponding element of the present architecture, and the evidence reported in this paper.}
\label{tab:mapping}
\centering
\footnotesize
\begin{tabular}{p{0.31\textwidth}p{0.31\textwidth}p{0.31\textwidth}}
\toprule
Demonstrated scope in prior work & Element of the present architecture & Evidence in this paper \\
\midrule
Latent denoising conditioned on an encoded strain-energy field \cite{herron2024}; penalties for floating material, insufficient material at load sites, and volume-fraction deviation added to the autoencoder loss \cite{lutheran2025}. & The topology prior is retained unchanged; a domain branch conditions denoising and a separately trained regressor steers the latent trajectory at sampling time. & Direct scores of decoded designs with and without guidance on training and held-out domains (Section~\ref{sec:results-guided}). \\
\addlinespace
Guidance of a graph-based latent diffusion over vector-quantized unit cells toward elastic targets with a separate property predictor \cite{yu2026}; editing of a partially noised reference design with a characteristic loss on the estimated clean latent \cite{chen2026}. & A regressor predicts the process-dependent utilization score from the noisy design latent, the encoded domain mask, and the process setting. & Fixed-geometry audit of the process-dependent score across seven settings (Section~\ref{sec:results-audit}). \\
\addlinespace
ControlNet spatial control \cite{zhang2023controlnet} combined with timestep-aware regressor guidance in image synthesis \cite{erel2025}. & A ControlNet domain branch and a regressor without a diffusion-time input, applied to a VQ-VAE-derived layout representation with a porous-design evaluator. & Guidance-weight sweep, start of guidance, and regressor accuracy by noise level (Section~\ref{sec:results-ablation}). \\
\bottomrule
\end{tabular}
\end{table*}

\subsection{Generative Machine Learning and Agentic AI for Topology Optimization} \label{sec:lit-gen}

TopoDiff \cite{maze2023} generates optimized topologies with a conditional diffusion model in image space, following earlier adversarial generators \cite{nie2021}, and trajectory alignment \cite{giannone2023} matches the sampling trajectory to intermediate iterates of a physics-based optimizer during training. Latent approaches \cite{herron2024,lutheran2025,yu2026,chen2026} differ in where engineering information enters: during representation learning, as a condition on denoising, or as a gradient during sampling (Table~\ref{tab:mapping}). The distinction matters here because, during sampling, only the score regressor receives the process setting, and its gradient steers denoising. Latent generation also introduces a reconstruction step that can alter thin features or narrow gaps, which matters when the objective depends on them.

Guidance adds objectives at sampling time. Classifier guidance shifts the mean of each reverse step along the gradient of a noise-aware classifier \cite{dhariwal2021}, and TopoDiff applies it with a compliance regressor and a floating-material classifier \cite{maze2023}. Closer to porous design, MicroLad \cite{lee2026} generates three-dimensional microstructures with a latent diffusion model trained on two-dimensional images and evaluates transport by computing relative diffusivity on the generated volumes, and Phan et al. \cite{phan2026} condition microstructure generation on processing variables such as annealing temperature and time. Connectivity of the solid phase has been addressed with differentiable constraints \cite{yang2026} and with learned rewards that steer a pretrained generator away from floating material \cite{kim2026}. Outside engineering design, PractiLight \cite{erel2025} combines ControlNet spatial control \cite{zhang2023controlnet} with a separately trained, timestep-aware regressor to control lighting in an image model.

Generative and analysis tools are also being placed inside agentic workflows, in which an LLM-based agent plans a task, calls tools, inspects the results, and revises its plan \cite{wang2024agents}. In design, a multi-agent framework has interpolated car shapes and predicted their aerodynamic performance with a surrogate \cite{elrefaie2025}, and a language-guided framework for metamaterials samples geometry-generator parameters with a conditional diffusion model and refines candidates with a surrogate \cite{zheng2026}; in additive manufacturing, an agent has evaluated alloys with thermophysical calculations and lack-of-fusion process maps \cite{pak2026}.

\begin{figure*}[t]\vspace*{4pt}
\centering
\setlength{\unitlength}{\textwidth}%
\begin{picture}(1,0.49294)
\put(0,0.39454){\includegraphics[width=\unitlength,trim=0 230.5bp 0 0,clip]{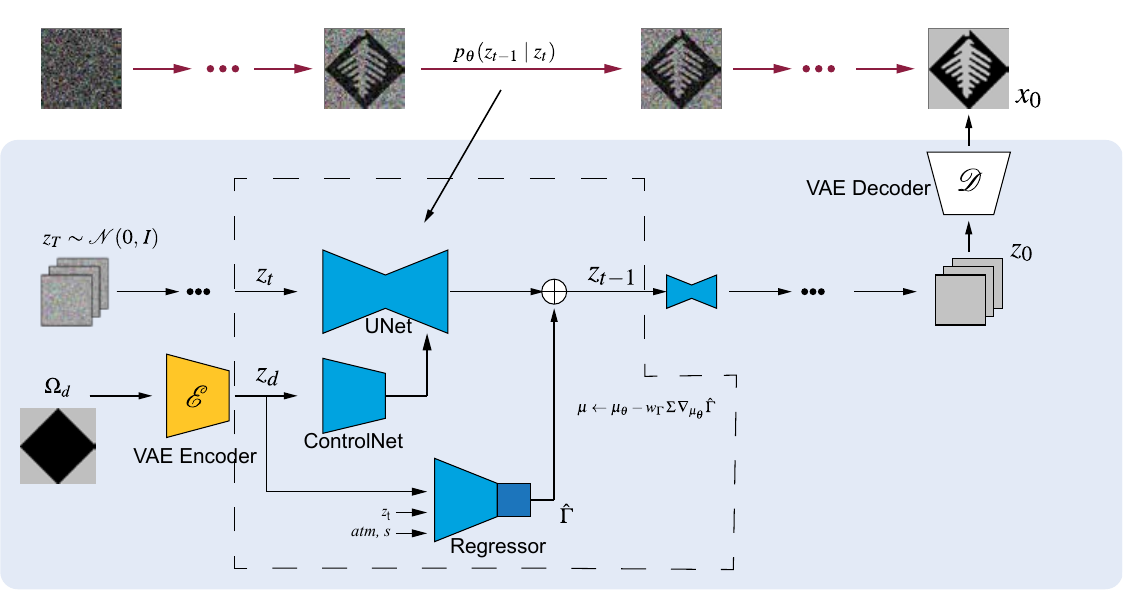}}
\put(0,0){\includegraphics[width=\unitlength]{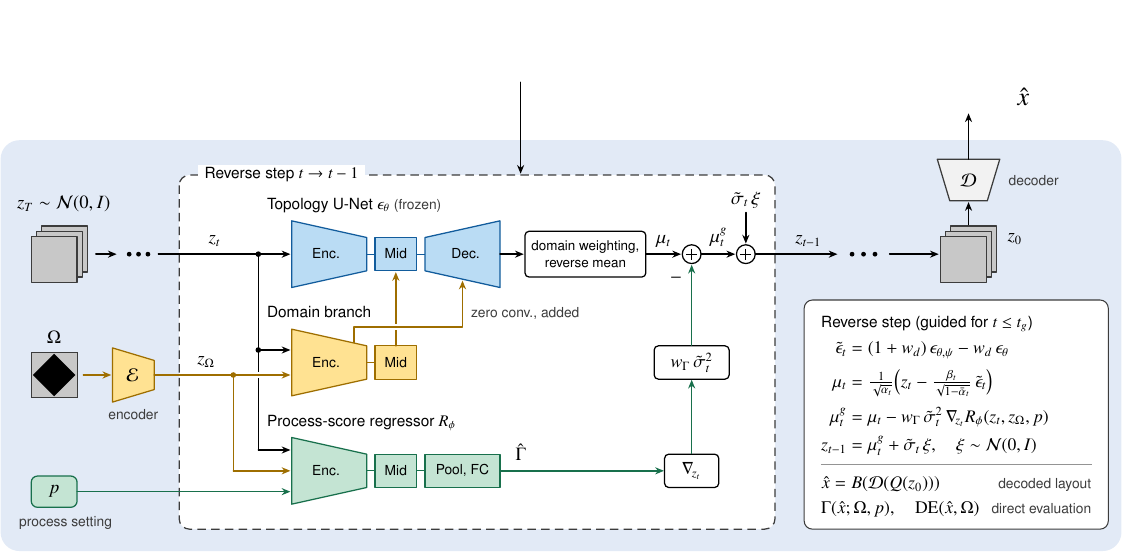}}
\end{picture}
\caption{Overview of the method. Top: schematic of reverse diffusion from noise to the decoded layout, in which black marks printed material. Bottom: one reverse step. The encoded domain mask $z_\Omega$ enters the domain branch, whose outputs are added to the middle block and the decoder of the topology U-Net through zero-initialized convolutions, and the noise predictions with and without the branch are weighted and converted to the reverse mean $\mu_t$. The process-score regressor receives the noisy latent, the domain latent, and the process setting $p$, and its gradient shifts the mean (Eqs.~(\ref{eq:cfg})--(\ref{eq:guided})). The U-Net and the domain branch also receive the embedding of the diffusion time; the regressor does not. The decoded layout is evaluated directly with the design score $\Gamma$ and the domain error DE.}
\label{fig:gr001}
\end{figure*}

\subsection{Limitations of Existing Approaches and Problem Statement} \label{sec:lit-gap}

Table~\ref{tab:mapping} relates the closest prior architectures to the elements of the present one and to the evidence reported in this paper. In those architectures, the conditions and guidance targets are structural, elastic, or image attributes. For the porous layouts studied here, the prescribed process setting changes the score used both to guide generation and to evaluate the decoded geometry. For a prescribed domain, does changing the process setting change which of the same candidate layouts receives the lowest score? With domain conditioning held fixed, how does the guidance weight affect the score recomputed on decoded layouts of training and held-out domains, relative to zero guidance? Section~\ref{sec:method} defines the score, the sampler, and the direct evaluator used to examine these questions, and Section~\ref{sec:case} applies them to AJP.

\section{Methodology} \label{sec:method}

Figure~\ref{fig:gr001} gives an overview of the method. A domain mask and a process setting are the inputs. A domain-conditioned latent diffusion sampler generates the layout, the gradient of a process-score regressor shifts each guided reverse step, and the decoded layout is evaluated directly. The method is formulated for a general process setting, and Section~\ref{sec:case} specifies it for aerosol jet printing.

\subsection{Design Problem and Manufacturing Constraints} \label{sec:problem}

A design is a binary layout $x \in \{0,1\}^{H \times W}$, in which $x_i = 1$ marks a pixel of printed material and $x_i = 0$ an unprinted pixel. The admissible region is a binary domain mask $\Omega \in \{0,1\}^{H \times W}$, and the unprinted pixels inside the domain form the channels. The process setting is a vector $p \in \mathcal{P} \subset \mathbb{R}^m$ of the process parameters that shape the deposit, where $\mathcal{P}$ is the operating range. The domain and the process setting are supplied as inputs, and the layout is the only optimized variable.

The functional score is the complement of the fraction of the domain occupied by printed material, with each printed pixel weighted by its proximity to the nearest channel. For a printed pixel $i$, let $d_i$ be the Euclidean distance to the nearest channel pixel, and let
\begin{equation}
u_i = \exp\!\left(-\frac{d_i}{L(p)}\right)
\label{eq:utility}
\end{equation}
be its utility, where $L(p) > 0$ is the utilization length, a modeled decay length: at a distance $L(p)$ from the nearest channel, the utility is $e^{-1}$. The process enters the score through $L(p)$ alone. A process model supplies this length for every setting in $\mathcal{P}$; the model may be an analytical relation, a calibration against printed samples, or a learned map. The design score is
\begin{equation}
\Gamma(x;\Omega,p) = 1 - \frac{\sum_i x_i\,\Omega_i\,u_i}{\sum_i \Omega_i}\,,
\label{eq:gamma}
\end{equation}
which lies in $[0,1]$ and is lower for better designs. With $f$ the printed fraction of the domain and $\bar{u}$ the mean utility of the printed pixels, Eq.~(\ref{eq:gamma}) reads $1 - \Gamma = f\,\bar{u}$, which separates a gain from added material from a gain from material lying closer to channels. Under Eq.~(\ref{eq:utility}), shortening $L(p)$ reduces the modeled utility of a printed pixel at a fixed positive distance from a channel. Layouts with different distributions of these distances can therefore change order when the process setting changes.

The score is a geometric proxy. It counts a printed pixel as served by the nearest channel pixel whether or not that channel connects to the open boundary, and it places no lower bound on the width of a feature. Containment is evaluated separately on the decoded layout. The domain error
\begin{equation}
\mathrm{DE}(x,\Omega) = \frac{\sum_i x_i\,(1-\Omega_i)}{\sum_i \Omega_i}\,,
\label{eq:de}
\end{equation}
reported as a percentage of the domain area, measures printed material outside the domain. The design problem addressed here is
\begin{equation}
\min_{x}\ \Gamma(x;\Omega,p) \quad \text{subject to} \quad \mathrm{DE}(x,\Omega) \le \delta ,
\label{eq:problem}
\end{equation}
where $\delta$ is a tolerance on the domain error. Only $\Gamma$ guides sampling; the domain branch of Section~\ref{sec:ldm} encourages containment, which is checked on the decoded layout.

\subsection{VQ-VAE Representation and Domain-Conditioned Latent Diffusion} \label{sec:ldm}

A VQ-VAE \cite{oord2017} with residual blocks consists of an encoder $\mathcal{E}$, a quantizer $Q$ that replaces each latent vector by its nearest codebook vector, and a decoder $\mathcal{D}$. For a layout $x$ with encoder output $z_e = \mathcal{E}(x)$ and selected codebook vectors $e = Q(z_e)$, it is trained with reconstruction, codebook, and commitment terms,
\begin{equation}
\mathcal{L}_{\mathrm{VQ}} = \lVert x - \mathcal{D}(e) \rVert^2 + \lVert \mathrm{sg}[z_e] - e \rVert^2 + \beta_c\, \lVert z_e - \mathrm{sg}[e] \rVert^2 ,
\label{eq:vq}
\end{equation}
where $\mathrm{sg}[\cdot]$ is the stop-gradient operator and $\beta_c$ is the commitment weight. Diffusion acts on the continuous encoder output before quantization, $z_0 = \mathcal{E}(x) \in \mathbb{R}^{c \times h \times w}$, and quantization is applied on the decoding path, so that a latent is decoded to the layout $\hat{x} = B(\mathcal{D}(Q(z_0)))$, where $B$ thresholds the decoder output, to which no sigmoid is applied, at 0.5. The forward process adds Gaussian noise over $T$ steps,
\begin{equation}
z_t = \sqrt{\bar\alpha_t}\, z_0 + \sqrt{1-\bar\alpha_t}\, \epsilon, \qquad \epsilon \sim \mathcal{N}(0, I),
\label{eq:forward}
\end{equation}
with $\bar\alpha_t = \prod_{s \le t} (1-\beta_s)$, so that $\sigma_t = \sqrt{1-\bar\alpha_t}$ is the noise level at step $t$. A U-Net $\epsilon_\theta(z_t, t)$ with residual and attention blocks is trained on the layout latents with the noise-prediction loss \cite{ho2020},
\begin{equation}
\mathcal{L}_{\mathrm{prior}}(\theta) = \mathbb{E}_{z_0,\,t,\,\epsilon} \left[ \lVert \epsilon - \epsilon_\theta(z_t, t) \rVert^2 \right].
\label{eq:prior}
\end{equation}
This network is the topology prior: it learns the distribution of porous layouts without reference to a domain or a process setting.

Domain control follows ControlNet \cite{zhang2023controlnet}. The domain mask is encoded with the same encoder, $z_\Omega = \mathcal{E}(\Omega)$, and passed to a trainable copy of the U-Net encoder and middle block. The outputs of this branch are added to the middle block and to the decoder of the U-Net through zero-initialized convolutions, so that training starts from the unmodified prior. With the branch attached, the network gives the conditional prediction $\epsilon_{\theta,\psi}(z_t, z_\Omega, t)$, where $\psi$ denotes the parameters of the branch and of its zero convolutions. They are trained with the same loss,
\begin{equation}
\mathcal{L}_{\mathrm{dom}}(\psi) = \mathbb{E}_{(z_0,\,z_\Omega),\,t,\,\epsilon} \left[ \lVert \epsilon - \epsilon_{\theta,\psi}(z_t, z_\Omega, t) \rVert^2 \right],
\label{eq:dom}
\end{equation}
while $\theta$ is kept fixed. At sampling time the conditional and unconditional noise predictions are combined with a domain weight $w_d$,
\begin{equation}
\tilde\epsilon_t = (1 + w_d)\, \epsilon_{\theta,\psi}(z_t, z_\Omega, t) - w_d\, \epsilon_\theta(z_t, t),
\label{eq:cfg}
\end{equation}
which gives the mean of the reverse step
\begin{equation}
\mu_t = \frac{1}{\sqrt{\alpha_t}} \left( z_t - \frac{\beta_t}{\sqrt{1-\bar\alpha_t}}\, \tilde\epsilon_t \right)
\label{eq:mean}
\end{equation}
with $\alpha_t = 1 - \beta_t$. Sampling $z_{t-1} = \mu_t + \tilde\sigma_t \xi$ with $\xi \sim \mathcal{N}(0,I)$ and $\tilde\sigma_t^2 = \beta_t (1-\bar\alpha_{t-1})/(1-\bar\alpha_t)$ defines the domain-conditioned sampler, which is the reference against which guidance is assessed.

\subsection{Manufacturability Assessment and Guidance} \label{sec:guidance}

The process-score regressor $R_\phi(z_t, z_\Omega, p)$ predicts the design score of the layout encoded in a noisy latent. It uses the encoder and middle block of the U-Net architecture with the design and domain latents as inputs, injects embeddings of the process parameters in the way the U-Net injects the embedding of the diffusion time, reduces the middle-block output by averaging over the spatial dimensions, and maps the result to a scalar with fully connected layers. Its parameters are initialized independently of the U-Net, and it has no diffusion-time input. Training pairs each layout with process settings drawn from $\mathcal{P}$, labels the pair with $\Gamma$ from Eq.~(\ref{eq:gamma}), and noises the latent to a random step with Eq.~(\ref{eq:forward}):
\begin{equation}
\mathcal{L}_{\mathrm{reg}}(\phi) = \mathbb{E}_{(x,\Omega,p),\,t,\,\epsilon} \left[ \left( \Gamma(x;\Omega,p) - R_\phi(z_t, z_\Omega, p) \right)^2 \right].
\label{eq:regloss}
\end{equation}
The regressor thus learns to predict the score of the clean layout from latents at all noise levels without being given the level. The same geometry appears with different labels at different process settings, so the regressor must learn the dependence of the score on the process as well as on the layout. The process model supplies the training targets of the regressor. During sampling, the regressor receives the process setting $p$, and its latent gradient shifts the reverse mean. The autoencoder, the topology prior, and the domain branch do not receive $p$.

We use the scaled negative gradient of the noisy-state score regressor as a local steering term in the reverse update \cite{dhariwal2021,maze2023}:
\begin{equation}
\mu_t^{g} = \mu_t - w_\Gamma\, \tilde\sigma_t^2\, \nabla_{z_t} R_\phi(z_t, z_\Omega, p), \qquad t \le t_g,
\label{eq:guided}
\end{equation}
followed by $z_{t-1} = \mu_t^{g} + \tilde\sigma_t \xi$. The gradient is evaluated at the current latent and is not normalized. The effective magnitude of the steering term depends on the score normalization, the latent gradient, and the sampler variance; the numerical weight $w_\Gamma$ is not a mixing fraction. We report it through the ratio $\rho_t = w_\Gamma \tilde\sigma_t \lVert \nabla_{z_t} R_\phi \rVert / \lVert \xi \rVert$ between the guidance displacement and the noise of the step, defined for $1 < t \le t_g$, the steps at which both guidance and noise are applied. Guidance is applied for $t \le t_g$, in the last part of sampling, where the noise level is low. Setting $w_\Gamma = 0$ in Eq.~(\ref{eq:guided}) gives the domain-conditioned sampler. Algorithm~\ref{alg:sampling} summarizes the procedure.

\begin{algorithm}[t]
\caption{Guided sampling and evaluation}
\label{alg:sampling}
\begin{algorithmic}[1]
\Require domain $\Omega$, process setting $p$, weights $w_d$, $w_\Gamma$, guidance start $t_g$, seed
\State $z_\Omega \gets \mathcal{E}(\Omega)$; \quad $z_T \sim \mathcal{N}(0, I)$
\For{$t = T, \dots, 1$}
  \State $\tilde\epsilon_t \gets (1+w_d)\,\epsilon_{\theta,\psi}(z_t, z_\Omega, t) - w_d\,\epsilon_\theta(z_t, t)$
  \State $\mu_t \gets \big(z_t - \beta_t \tilde\epsilon_t / \sqrt{1-\bar\alpha_t}\big)/\sqrt{\alpha_t}$
  \If{$t \le t_g$}
    \State $\mu_t \gets \mu_t - w_\Gamma \tilde\sigma_t^2\, \nabla_{z_t} R_\phi(z_t, z_\Omega, p)$
  \EndIf
  \State $\xi \sim \mathcal{N}(0,I)$ if $t > 1$, else $\xi \gets 0$
  \State $z_{t-1} \gets \mu_t + \tilde\sigma_t \xi$
\EndFor
\State $\hat{x} \gets B(\mathcal{D}(Q(z_0)))$
\State \Return $\hat{x}$, $\Gamma(\hat{x};\Omega,p)$, $\mathrm{DE}(\hat{x},\Omega)$
\end{algorithmic}
\end{algorithm}

\subsection{Independent Design Evaluation} \label{sec:eval}

Performance is measured on the decoded layouts with the direct evaluator, and regressor predictions are not used for reporting it. The design score $\Gamma(\hat{x};\Omega,p)$ of Eq.~(\ref{eq:gamma}) measures the modeled utilization, and the domain error $\mathrm{DE}(\hat{x},\Omega)$ of Eq.~(\ref{eq:de}) measures containment. Guided sampling is assessed against the sampler with $w_\Gamma = 0$: the mean score at each guidance weight is compared with the mean at $w_\Gamma = 0$ over the same domains and process settings, and a negative change indicates that guidance lowered the score. The decoded layouts are evaluated without clipping to the domain and without repair. The accuracy of the regressor is reported as the coefficient of determination $R^2$ between its predictions on noised latents and the direct scores, by noise level. The trajectory curves shown in Section~\ref{sec:case} are regressor predictions for the clean latent estimated at each step; they are a diagnostic, distinct from the evaluation of the regressor at the noisy latent in Eq.~(\ref{eq:guided}), and are reported separately from the direct scores of the final layouts.

The dependence of the score on the process is examined before any generation by a fixed-geometry audit. A set of candidate layouts in one domain is evaluated at several process settings with Eq.~(\ref{eq:gamma}), and the lowest-scoring candidate is recorded at each setting. A common positive factor, $\Gamma(x;p) = a(p)\,\Gamma_0(x)$, cannot change the ranking within a fixed candidate pool. A change of the lowest-scoring candidate across settings therefore shows that the dependence of the score on the process is geometry-dependent in that pool.

\section{Case study} \label{sec:case}

\subsection{Porous Architecture Design Task and Data} \label{sec:data}

\begin{figure*}[t]\vspace*{4pt}
\centerline{\includegraphics[width=0.92\textwidth]{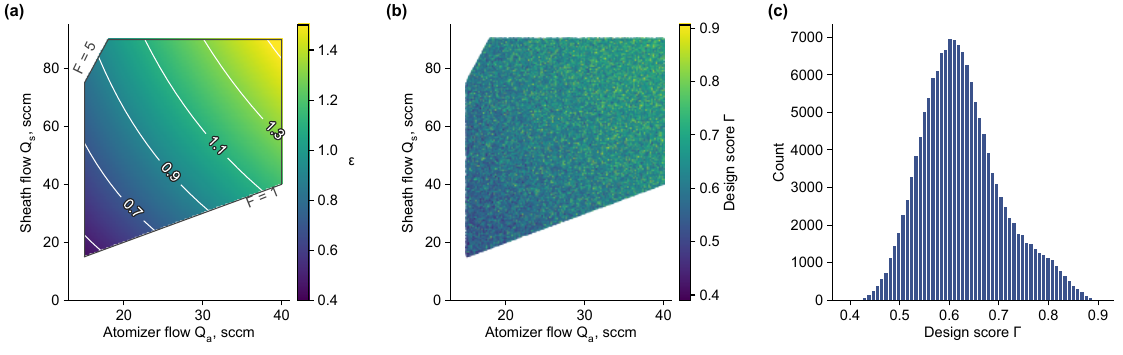}}
\caption{Process map and score data. (a) Process--morphology factor $\varepsilon$ over the modeled operating range bounded by focusing ratios of 1 and 5. (b) Design score $\Gamma$ of the regressor data over the same range. (c) Histogram of $\Gamma$ in the regressor data.}
\label{fig:gr002}
\end{figure*}

The method is applied to planar porous layouts for aerosol jet printing, a process with which lithium iron phosphate cathodes with a bimodal pore distribution have been printed \cite{rodriguez2021}. The process setting is $p = (Q_a, Q_s)$, the atomizer and sheath gas flow rates in standard cubic centimeters per minute (sccm), so that $m = 2$. The focusing ratio $F = Q_s / Q_a$ sets the diameter $D$ of the focused aerosol stream relative to the nozzle diameter $D_n$ \cite{secor2018,phuah2020},
\begin{equation}
\frac{D(F)}{D_n} = \sqrt{1 - \sqrt{\frac{F}{1+F}}}\,,
\label{eq:focus}
\end{equation}
and the atomizer flow influences aerosol delivery. The two are combined in a dimensionless process--morphology factor
\begin{equation}
\varepsilon(p) = \mathrm{clip}\!\left( \frac{Q_a}{Q_{a,0}} \, \frac{D(F_0)}{D(F)},\ 0.4,\ 1.5 \right),
\label{eq:eps}
\end{equation}
which equals one at the reference setting $p_0 = (30, 45)$~sccm with $F_0 = 1.5$ and is limited to the interval $[0.4, 1.5]$, and the utilization length of Eq.~(\ref{eq:utility}) is modeled as
\begin{equation}
L(p) = \frac{L_0}{\varepsilon(p)}\,,
\label{eq:Lp}
\end{equation}
where $L_0$ is its value at the reference setting. The factor grows with the material delivered by the atomizer flow and with the focusing of the stream, in line with the reported dependence of printed line morphology on the two flows \cite{mahajan2013,phuah2020}; neither term is read as a line width or a line thickness on its own. Equation~(\ref{eq:Lp}) is a modeling assumption that assigns a shorter utilization length to settings with a larger factor. The reference atomizer flow of 30~sccm is the operating setting reported for an Optomec Aerosol Jet 5X system with a 150~$\mu$m nozzle \cite{nguyen2026thesis}. The modeled operating range $\mathcal{P}$ is $1 \le F \le 5$ with $15 \le Q_a \le 40$~sccm and $Q_s \le 90$~sccm (Fig.~\ref{fig:gr002}(a)).

A layout occupies a $128 \times 128$ grid that represents a design space of 10~mm $\times$ 10~mm, so that a pixel measures about 78~$\mu$m. The printed phase is the functional material, and the unprinted pixels inside the domain are the channels. The utilization length at the reference setting is $L_0 = 8$ pixels, half of the nominal minimum wall width of 16 pixels used in constructing the layouts; over the modeled range, $L(p)$ ranges from about 5.3 to 18.2 pixels.

\begin{figure*}[t]\vspace*{4pt}
\centerline{\includegraphics[width=0.92\textwidth]{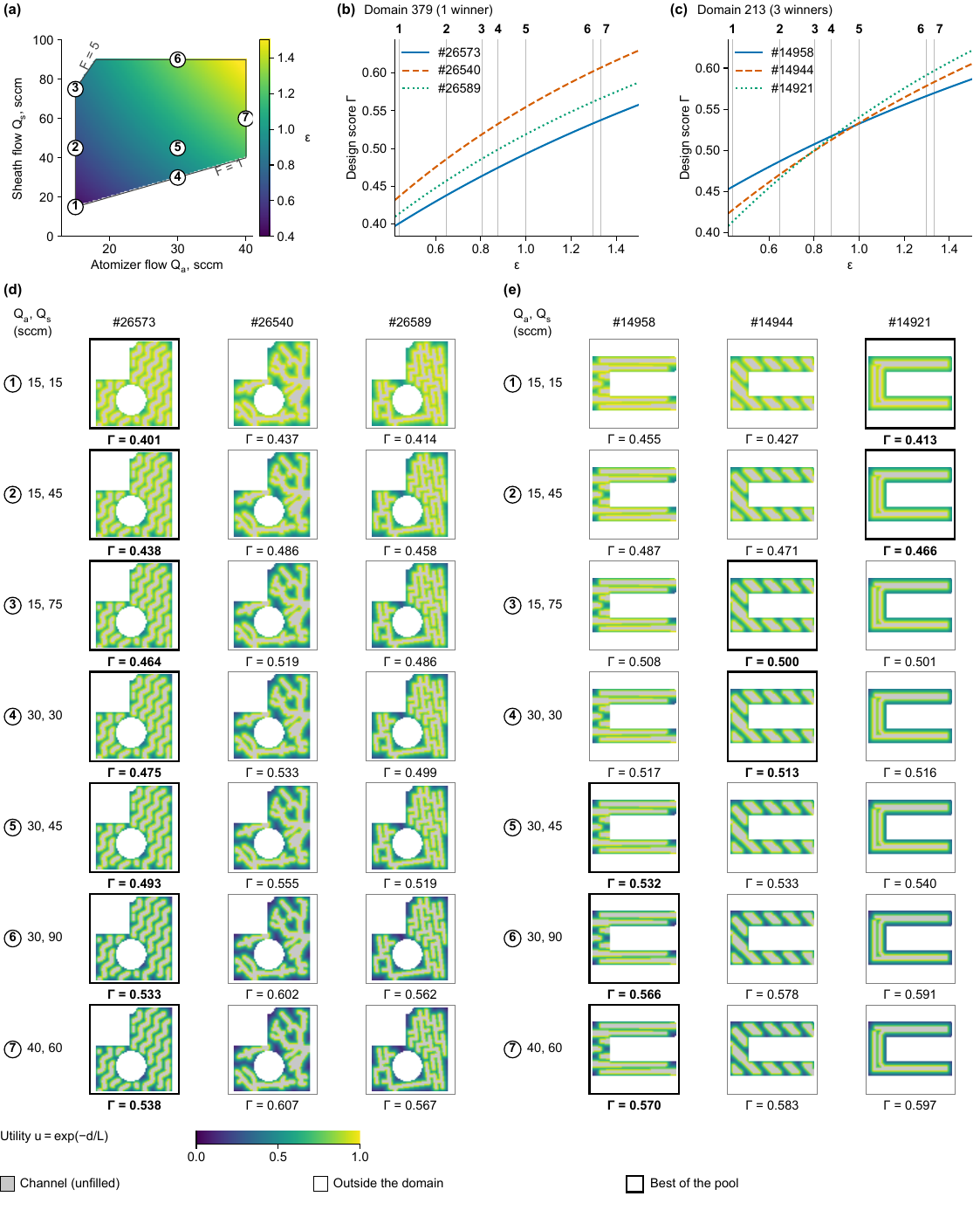}}
\caption{Fixed-geometry audit of the design score. (a) The seven test settings on the map of $\varepsilon$. (b) Scores of the three displayed layouts of Domain~379 as functions of $\varepsilon$; one layout is the lowest at all settings. (c) Scores of the three displayed layouts of Domain~213; the lowest-scoring layout changes twice. (d), (e) Utility maps and scores of these layouts at the seven settings for Domain~379 and Domain~213. Colors show the utility of the printed pixels, gray marks channels, white lies outside the domain, and the black frame marks the lowest score in each row.}
\label{fig:gr003}
\end{figure*}

\begin{figure}[t]\vspace*{4pt}
\centerline{\includegraphics[width=0.95\columnwidth]{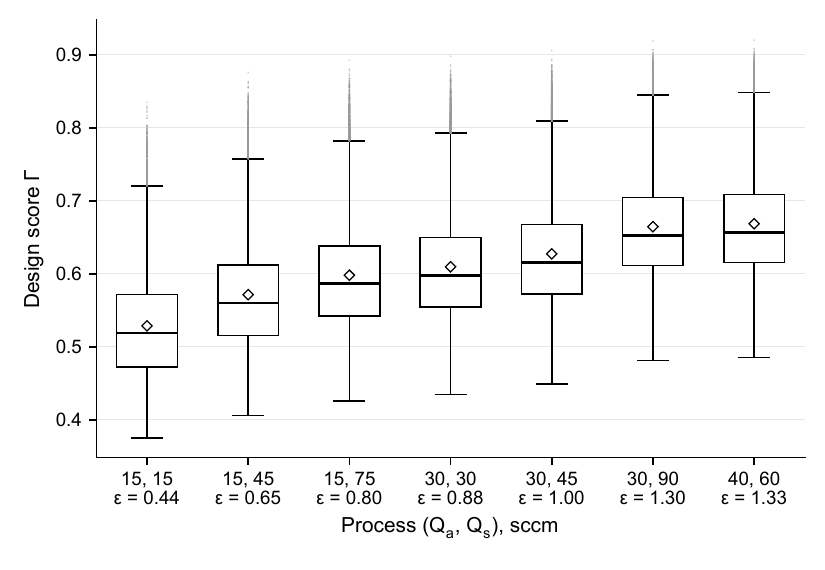}}
\caption{Distribution of the design score $\Gamma$ over the layouts of the corpus at the seven test settings. Boxes span the quartiles, lines mark medians, and diamonds mark means.}
\label{fig:gr004}
\end{figure}

The layout corpus contains 500 domains with the outlines of engineering parts, such as rectangles, discs, I-sections, lugs, and notched plates, each combined with about 70 channel patterns, which gives 34,991 layouts. The domains are divided into 450 for training, 38 for validation, and 12 for testing. Validation data were used for model selection and not for gradient updates. The regressor data are 150,000 records, each a layout paired with a process setting drawn at random from $\mathcal{P}$ and labeled with the score of Eq.~(\ref{eq:gamma}), divided in similar proportions. Figure~\ref{fig:gr002}(b) and (c) show the resulting scores, which range from about 0.4 to about 0.9 with a mode near 0.6. Seven process settings, marked in Fig.~\ref{fig:gr003}(a) and listed in Table~\ref{tab:audit}, are used for testing; they cover $\varepsilon$ from 0.44 to 1.33. Figure~\ref{fig:gr004} shows the distribution of the score over all layouts of the corpus at each of these settings; the median rises from about 0.52 at $\varepsilon = 0.44$ to about 0.66 at $\varepsilon = 1.33$.

\subsection{Implementation Verification and Experimental Protocol} \label{sec:protocol}

The implementation follows the latent diffusion model of \cite{rombach2022}. The VQ-VAE uses residual blocks without attention, a codebook of 1,024 vectors, of which 253 are in use after training, and $\beta_c = 0.25$. It maps a layout to a $4 \times 32 \times 32$ latent. Diffusion operates on latents multiplied by a constant scale that is computed on the training latents and used again at sampling; in the equations, the scale is included in the encoding and decoding operations. The U-Net has a base width of 96 channels with multipliers $(1, 2, 4)$, and the regressor has the same encoder and middle-block layout followed by a fully connected layer of 768 units. The autoencoder and the U-Net have more residual blocks in the decoding path than in the encoding path.

All networks are trained with a learning rate of $10^{-4}$ and a cosine schedule, and the codebook is updated with a rate of $10^{-2}$. The autoencoder is trained with Eq.~(\ref{eq:vq}) and the U-Net with Eq.~(\ref{eq:prior}) on the layout latents; the domain branch is then trained with Eq.~(\ref{eq:dom}) with the U-Net frozen, and the regressor is trained separately on noised latents with Eq.~(\ref{eq:regloss}).

Sampling uses $T = 1{,}000$ steps with the noise schedule of \cite{rombach2022}, in which $\sqrt{\beta_t}$ increases linearly from $\sqrt{0.0015}$ to $\sqrt{0.0195}$, and a domain weight $w_d = 3$. Guidance starts at $t_g = 250$, where the noise level is $\sigma_t = 0.70$ and the signal-to-noise ratio $\bar\alpha_t/(1-\bar\alpha_t)$ is 1.04. The implementation and the figures index the steps from 999 to 0, so that guidance covers the indices 249 to 0. The tested weights are $w_\Gamma \in \{0,\ 300,\ 8{,}000,\ 12{,}000\}$, where $w_\Gamma = 0$ is the domain-conditioned sampler.

The sampling study covers ten domains, six from the training set (11, 79, 164, 210, 286, and 379) and four held out from training, three from the validation set (462, 464, and 485) and one from the test set (499), at the seven process settings, with four samples for each domain, setting, and weight. This gives 280 scores per weight and 1,120 in total. For a batch of four layouts on an NVIDIA A100 graphics processing unit, sampling took about 58~s with domain conditioning alone and about 61~s with guidance.

\begin{table}[t]
\caption{Design score $\Gamma$ of the layouts displayed in Fig.~\ref{fig:gr003} at the seven process settings. The lowest score in each domain and setting is in bold.}
\label{tab:audit}
\centering
\footnotesize
\setlength{\tabcolsep}{3pt}
\begin{tabular}{@{}ccccccccc@{}}
\toprule
 & & \multicolumn{3}{c}{Domain 379} & & \multicolumn{3}{c}{Domain 213} \\
\cmidrule(lr){3-5}\cmidrule(l){7-9}
$(Q_a, Q_s)$ & $\varepsilon$ & \#26573 & \#26540 & \#26589 & & \#14958 & \#14944 & \#14921 \\
\midrule
(15, 15) & 0.44 & \textbf{0.401} & 0.437 & 0.414 & & 0.455 & 0.427 & \textbf{0.413} \\
(15, 45) & 0.65 & \textbf{0.438} & 0.486 & 0.458 & & 0.487 & 0.471 & \textbf{0.466} \\
(15, 75) & 0.80 & \textbf{0.464} & 0.519 & 0.486 & & 0.508 & \textbf{0.500} & 0.501 \\
(30, 30) & 0.88 & \textbf{0.475} & 0.533 & 0.499 & & 0.517 & \textbf{0.513} & 0.516 \\
(30, 45) & 1.00 & \textbf{0.493} & 0.555 & 0.519 & & \textbf{0.532} & 0.533 & 0.540 \\
(30, 90) & 1.30 & \textbf{0.533} & 0.602 & 0.562 & & \textbf{0.566} & 0.578 & 0.591 \\
(40, 60) & 1.33 & \textbf{0.538} & 0.607 & 0.567 & & \textbf{0.570} & 0.583 & 0.597 \\
\bottomrule
\end{tabular}
\end{table}

\begin{figure*}[t]\vspace*{4pt}
\centerline{\includegraphics[width=0.95\textwidth]{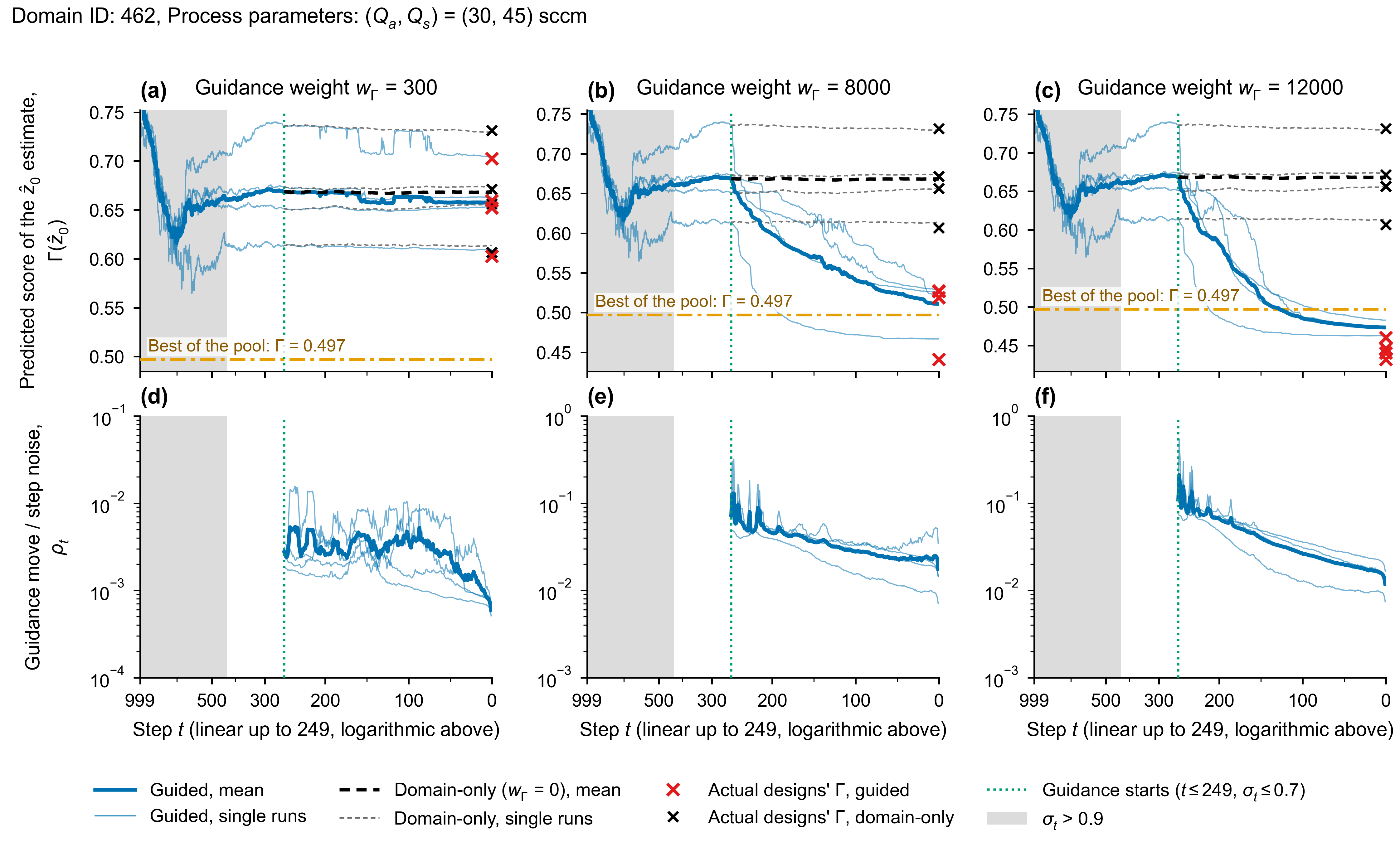}}
\caption{Sampling runs for validation Domain~462 at $(Q_a, Q_s) = (30, 45)$~sccm at guidance weights of 300, 8,000, and 12,000 with $w_d = 3$. Top: regressor prediction for the clean latent $\hat{z}_0$ estimated at each step from $z_t$ and the predicted noise, with crosses marking the directly computed scores of the decoded designs. Bottom: ratio $\rho_t$ of the guidance displacement to the noise of the step. Thin lines are individual runs, thick lines are means, and dashed lines are the domain-conditioned sampler with the same noise. The horizontal line marks the lowest score among the layouts of this domain in the corpus. The dotted line marks the start of guidance, and the shaded band marks noise levels above 0.9; the step axis uses the index of the implementation, from 999 to 0, and is logarithmic above 249. Vertical scales differ across panels.}
\label{fig:trace}
\end{figure*}

\subsection{Feasibility and Design Performance} \label{sec:results}

\subsubsection{Process dependence of the preferred layout} \label{sec:results-audit}

The fixed-geometry audit of Section~\ref{sec:eval} is applied to the layouts of each domain, about 70 per domain, at the seven process settings. In 309 of the 500 domains, a single layout has the lowest score at all seven settings; in 177 domains, two different layouts have the lowest score at different settings, and in 14 domains, three. The counts above use unrounded scores. With the comparison tolerance set to 0.01, the audit returned a single lowest-scoring layout in 473 domains and two in 27.

Figure~\ref{fig:gr003} shows two domains, and Table~\ref{tab:audit} lists the scores of the displayed layouts. For Domain~213 the three displayed layouts are each the lowest-scoring one at some setting; for Domain~379 a single layout is the lowest at all seven settings, and the two layouts with the next-lowest mean scores are displayed with it. In Domain~379, layout \#26573 leads the displayed layouts by 0.013 at $\varepsilon = 0.44$ and by 0.029 at $\varepsilon = 1.33$, while its own score rises from 0.401 to 0.538. In Domain~213 the lowest-scoring layout changes twice across the seven settings. Layout \#14921, whose channels follow the outline of the domain between wide printed bands, is preferred at the two settings with the smallest $\varepsilon$; layout \#14944, with short diagonal channels, is preferred at the two intermediate settings; and layout \#14958, with long channels parallel to the arms of the domain, is preferred at the three settings with the largest $\varepsilon$. Layout \#14958 scored 0.042 higher than \#14921 at $(15, 15)$~sccm (0.455 against 0.413) and 0.027 lower at $(40, 60)$~sccm (0.570 against 0.597). Among the displayed layouts, the smallest winning margin in this domain was 0.001 at the reported precision.

The ranking changes within Domain~213, so in this domain the dependence of the score on the process is geometry-dependent. Domain~379 is the complementary case: the setting shifts the score of the preferred layout by 0.137 and leaves the choice of layout unchanged.

\subsubsection{Guided sampling} \label{sec:results-guided}

Figure~\ref{fig:trace} shows sampling runs for validation Domain~462 at $(Q_a, Q_s) = (30, 45)$~sccm, with four guided runs at each weight and four runs of the domain-conditioned sampler with the same noise. Before guidance starts, the guided and domain-conditioned prediction trajectories coincide. At the end of sampling, the mean predictions at weights of 300, 8,000, and 12,000 are 0.658, 0.511, and 0.473, against 0.668 with domain conditioning alone.

The directly computed scores of the decoded designs lie between 0.606 and 0.731 with domain conditioning alone, between 0.603 and 0.702 at a weight of 300, between 0.441 and 0.527 at 8,000, and between 0.432 and 0.460 at 12,000. In the four matched runs at each weight, the guided design scores lower than its counterpart generated with the same noise and zero guidance, by 0.004 to 0.029 at a weight of 300, by 0.130 to 0.212 at 8,000, and by 0.175 to 0.271 at 12,000. The lowest score among the layouts of this domain in the corpus is 0.497 at this setting; the four designs at 12,000 and one design at 8,000 score lower.

\begin{figure*}[t]\vspace*{4pt}
\centerline{\includegraphics[width=0.80\textwidth]{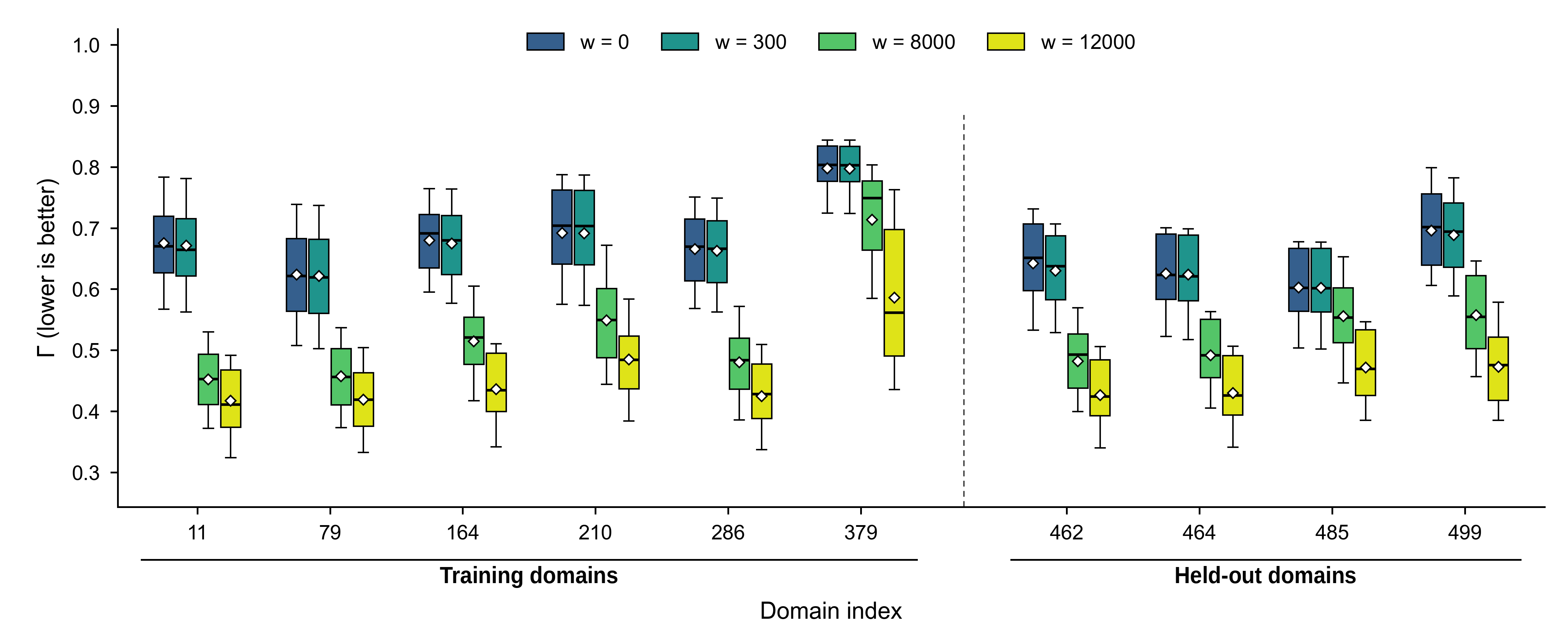}}
\caption{Directly computed score $\Gamma$ of the decoded designs in each domain of the sampling study at the four guidance weights, denoted $w$ in the legend. Each box covers 28 scores (seven settings and four samples); boxes span the quartiles, lines mark medians, and diamonds mark means. The held-out domains comprise three validation domains (462, 464, and 485) and one test domain (499).}
\label{fig:bydomain}
\end{figure*}

Table~\ref{tab:main} and Fig.~\ref{fig:bydomain} give the results of the sampling study. With domain conditioning alone, the mean directly computed score is 0.689 on the training domains and 0.642 on the held-out domains. Relative to these values, the mean changes at weights of 300, 8,000, and 12,000 are $-0.003$, $-0.161$, and $-0.228$ on the training domains and $-0.006$, $-0.120$, and $-0.192$ on the held-out domains. In each of the ten domains, the mean score at 8,000 and at 12,000 is lower than with domain conditioning alone, and the mean at 12,000 is lower than at 8,000. The interquartile range does not change uniformly across domains; in Domain~379 it is widest at 12,000.

\begin{table}[t]
\caption{Mean directly computed score $\Gamma$ of the decoded designs in the sampling study and its change relative to $w_\Gamma = 0$. Each mean covers seven settings and four samples per setting: 168 scores for the six training domains and 112 scores for the four held-out domains (three validation domains and one test domain).}
\label{tab:main}
\centering
\footnotesize
\setlength{\tabcolsep}{4pt}
\begin{tabular}{@{}lcccc@{}}
\toprule
 & \multicolumn{2}{c}{Training domains} & \multicolumn{2}{c}{Held-out domains} \\
\cmidrule(lr){2-3}\cmidrule(l){4-5}
Guidance weight $w_\Gamma$ & Mean $\Gamma$ & Change & Mean $\Gamma$ & Change \\
\midrule
0 (domain-conditioned) & 0.6889 & -- & 0.6416 & -- \\
300 & 0.6863 & $-0.0026$ & 0.6359 & $-0.0057$ \\
8,000 & 0.5276 & $-0.1613$ & 0.5215 & $-0.1201$ \\
12,000 & 0.4611 & $-0.2278$ & 0.4499 & $-0.1917$ \\
\bottomrule
\end{tabular}
\end{table}

\subsection{Ablation and Independent Validation} \label{sec:results-ablation}

Because $w_\Gamma$ and $w_d$ act on different terms of the update, their ratio does not measure the relative influence of the two paths. The bottom panels of Fig.~\ref{fig:trace} express the weights as displacements relative to the noise of the step. At $w_\Gamma = 300$ the mean ratio $\rho_t$ is a few thousandths; at $8{,}000$ and $12{,}000$ it is about 0.1 when guidance starts and about 0.02 and 0.01, respectively, near the end of sampling. These ratios are not fractions of the whole update. The regressor is less accurate at high noise: on the validation data, $R^2$ is 0.58 for $\sigma_t \ge 0.9$ and close to one at lower noise levels. Guidance is enabled for the final 250 steps, starting at a signal-to-noise ratio of about one, within the lower-noise region where the validation predictions are more accurate.

For the runs of Fig.~\ref{fig:trace}, the mean direct scores at the three weights are 0.654, 0.503, and 0.444, against 0.666 with domain conditioning alone; the corresponding mean predictions are 0.658, 0.511, and 0.473, against 0.668. The decrease that the regressor predicts is thus also found by the direct evaluator, and at the largest weight the direct scores are lower than the predictions by 0.023 to 0.033. For containment, the domain error of Eq.~(\ref{eq:de}) against the prescribed mask has a mean of 0.0033\% and a maximum of 0.0437\% over all 1,120 decoded designs, with or without guidance; 238 designs have at least one printed pixel outside the mask. With the mask dilated by a $5\times5$ square kernel for counting outside pixels and the denominator unchanged, no design had a printed pixel outside it. The decoded designs were not modified.

\section{Discussion} \label{sec:discussion}

\subsection{Manufacturing and Design Implications} \label{sec:implications}

The fixed-geometry audit shows two ways in which a process setting bears on a porous layout. In Domain~379, changing the setting changed the scores without changing the preferred layout. In Domain~213 the setting changes which layout is preferred, so a layout selected for one setting is not the layout to select for another. The audit uses only the direct evaluator and identifies the domains in which the modeled process setting affects layout selection.

The form of the score suggests the mechanism. A larger process--morphology factor shortens the utilization length, so printed material far from a channel contributes less and layouts with more closely spaced channels gain relative to layouts with wide printed bands. The order of preference in Domain~213 follows this pattern, from the layout with wide printed bands at small $\varepsilon$ to the layout with long parallel channels at large $\varepsilon$. The regressor receives the process setting so that this modeled preference can influence generation during sampling.

\subsection{Relationship to Prior Work} \label{sec:prior}

Each element of the architecture has a precedent (Table~\ref{tab:mapping}). Regressor guidance of a diffusion sampler toward an engineering objective is established in image space \cite{maze2023} and in latent space \cite{yu2026,chen2026}, and spatial control combined with a timestep-aware regressor has been demonstrated in image synthesis \cite{erel2025}. The present study applies this pattern to a task in which the objective depends on manufacturing process variables and the admissible region is an arbitrary domain. Its contribution lies in the process-dependent formulation, in the evidence that the formulation changes layout preference, and in the direct comparison of decoded designs with and without guidance on training and held-out domains.

The domain branch was trained with the topology prior frozen; the regressor was initialized independently and trained on noised layout latents, and only the regressor receives the process setting. Where the same layout representation and distance-based score suit another process, the architecture permits the autoencoder, the topology prior, and the domain branch to be held fixed while the regressor is trained on labels from the model of the utilization length for that process. The process map of Eqs.~(\ref{eq:focus})--(\ref{eq:Lp}) and the reported evidence are specific to aerosol jet printing.

\subsection{Limitations and Practical Use} \label{sec:limits}

The score ranks decoded layouts by modeled utilization. The process--morphology factor and its link to the utilization length have yet to be calibrated against printed deposits. If calibration changes $L(p)$ while the layout representation remains applicable, the direct evaluator can supply revised labels for retraining the regressor while the other networks remain fixed. Computing the score on the decoded layout accounts for the geometry actually returned by the decoder. Before fabrication, candidate layouts also require checks on the minimum widths of printed features and open channels and on channel access from the boundary. These checks were not part of the reported evaluation.

With the guidance start held fixed, the group means changed little at a weight of 300 and fell further at 8,000 and 12,000. Testing additional weights and start times would inform their selection. In Fig.~\ref{fig:trace}, for validation Domain~462 at a single process setting, all four designs at each of the two higher weights scored below the best of the four domain-conditioned designs. A comparison with selection from a larger domain-conditioned pool under the same time budget remains open.

\section{Concluding remarks and future work} \label{sec:conclusion}

This paper asked whether process knowledge can steer a latent diffusion generator toward porous layouts suited to a given process setting inside a prescribed domain. A process-dependent utilization score was formulated in which a process model sets the utilization length, and it was specified for aerosol jet printing from the atomizer flow and the flow-focusing relation. Evaluated at seven modeled process settings, the lowest-scoring candidate layout of a domain differed between settings in 191 of 500 domains, which shows that the dependence of the score on the process is geometry-dependent. A score regressor trained on noised latents, combined with a ControlNet domain branch on a frozen topology prior, was used to steer sampling. Over ten domains and seven settings, guidance at the largest tested weight lowered the mean directly computed score from 0.689 to 0.461 on six domains of the training set and from 0.642 to 0.450 on three validation domains and one test domain, and a weight of 300 left it within 0.006 of the value without guidance. These comparisons establish changes in a modeled utilization score; deposited geometry and device performance were not measured.

Three steps follow. The first is calibration of the process--morphology factor and of the minimum feature and channel widths against printed lines, so that the feasibility checks rest on measurements. The second is fabrication of generated designs at different process settings together with an independent transport evaluation of the same designs. The third is the inclusion of the feasibility criteria in the guidance target, after which three-dimensional layouts, additional process variables, and other deposition processes can be considered. For agentic design workflows, the sampler and the direct evaluator are the two components with which a design agent could request candidates for a process setting and check them. A next step is to connect these components to such an agent and to test whether feedback from the evaluator helps it request better candidates within a fixed sampling budget.

\section*{Declaration of generative AI and AI-assisted technologies in the manuscript preparation process}

During the preparation of this work the authors used Claude (Anthropic) and ChatGPT (OpenAI) to assist with editing and formatting the manuscript. The authors reviewed and edited the content as needed and take full responsibility for the final manuscript.

\bibliographystyle{elsarticle-harv-citeorder}
\bibliography{references_final}

\end{document}